\documentclass[runningheads]{llncs}
\usepackage[T1]{fontenc}
\usepackage{hyperref}
\usepackage{graphicx}
\usepackage{multirow}
\usepackage{subfigure}
\usepackage{amsfonts}
\usepackage{floatrow}
\begin{document}
\title{Verifiable and Energy Efficient Medical Image Analysis with Quantised Self-attentive Deep Neural Networks}
\titlerunning{Quantised Self-attentive Deep Neural Networks}
%
\author{Rakshith Sathish\inst{1} \and
Swanand Khare\inst{1} \and
Debdoot Sheet\inst{1}}
%
\institute{Indian Institute of Technology Kharagpur, West Bengal, India}
%
\maketitle              
\begin{abstract}
Convolutional Neural Networks have played a significant role in various medical imaging tasks like classification and segmentation. They provide state-of-the-art performance compared to classical image processing algorithms. However, the major downside of these methods is the high computational complexity, reliance on high-performance hardware like GPUs and the inherent black-box nature of the model. In this paper, we propose quantised stand-alone self-attention based models as an alternative to traditional CNNs. In the proposed class of networks, convolutional layers are replaced with stand-alone self-attention layers, and the network parameters are quantised after training. We experimentally validate the performance of our method on classification and segmentation tasks. We observe $50-80\%$ reduction in model size, $60-80\%$  lesser number of parameters, $40-85\%$ fewer FLOPs and $65-80\%$ more energy efficiency during inference on CPUs. The code will be available at \href {https://github.com/Rakshith2597/Quantised-Self-Attentive-Deep-Neural-Network}{https://github.com/Rakshith2597/Quantised-Self-Attentive-Deep-Neural-Network}. 

\end{abstract}

\section{Introduction}

Deep neural networks have played a significant role in medical image analysis. Since the advent of UNet\cite{ronneberger2015u} to UNetr\cite{hatamizadeh2022unetr}, the performance of neural networks on various tasks like classification, segmentation, and restoration has improved considerably. Deeper and broader convolutional neural networks generally show an improvement in performance at the cost of an increase in the number of learnable parameters, model size and total floating-point operations performed during a single forward pass of the data through the network. Moreover, these models require specialised high-performance hardware even during inference. This reliance on larger models and high-performance hardware hinders the last-mile delivery of AI solutions to improve the existing healthcare system, especially in resource constrained developing and under-developed countries.\\


\noindent\textbf{Challenges:} The performance and trustability of deep neural network-based methods are of utmost importance, especially in the medical domain. The performance of these methods decreases as we try to reduce the number of learnable parameters in the model. As an example, in the case of image classification, deeper networks have been shown to be superior to shallow networks with fewer parameters \cite{he2016deep,huang2017densely}. Despite the good performance measured in terms of quantitative evaluation metrics, deep neural network (DNN) are known to make the right decision for the wrong reasons \cite{chakravartyradiologist}. This limits the trustability of DNN-based frameworks in practical application. Additionally, the black box nature of the convolutional neural networks makes them unreliable for clinical applications. Developing a method that relies on fewer parameters and is clinically verifiable is a challenging task. Also, an efficient model is expected to replicate the performance during inference at a resonable execution speed even in the absence of GPUs. 

Attention-based networks were proposed to augment DNNs with explainability in the case of natural images. However, due to the inherent differences in the nature of images, we cannot assume an equivalent performance in the medical images. As an example, in detecting objects in natural images, the objects of interest often have a well-defined shape and structure, which are absent in the case of medical images. In the case of medical image classification, the biomarkers are usually unstructured pathologies with variable appearance. In this work, we try to verify the effectiveness of replacing convolutions with attention in neural networks for medical images.\\

\noindent\textbf{Related works:} Transformers \cite{vaswani2017attention}, based solely on attention mechanisms has revolutionised the way models are designed for natural language tasks. Motivated by their success, \cite{zhu2020deformable}, \cite{ye2019cross}, \cite{ramachandran2019stand} and \cite{yang2020learning} explored the possibility of using self-attention to solve various vision tasks. Among these, the stand-alone self-attention proposed by \cite{ramachandran2019stand} established that self-attention could potentially replace convolutional layers altogether. Even though it is efficient compared to other DNNs, such models can be further improved by quantising the weights and activations of the networks \cite{paupamah2020quantisation}. The quantisation of deep neural networks has shown significant progress in recent years \cite{xu2021mixed}\cite{aji2020compressing}. The ability to quantise the neural network trained in high precision without substantial loss in performance during inference simplifies the process. 

\begin{figure}
    \centering
    \includegraphics[width=0.8\linewidth]{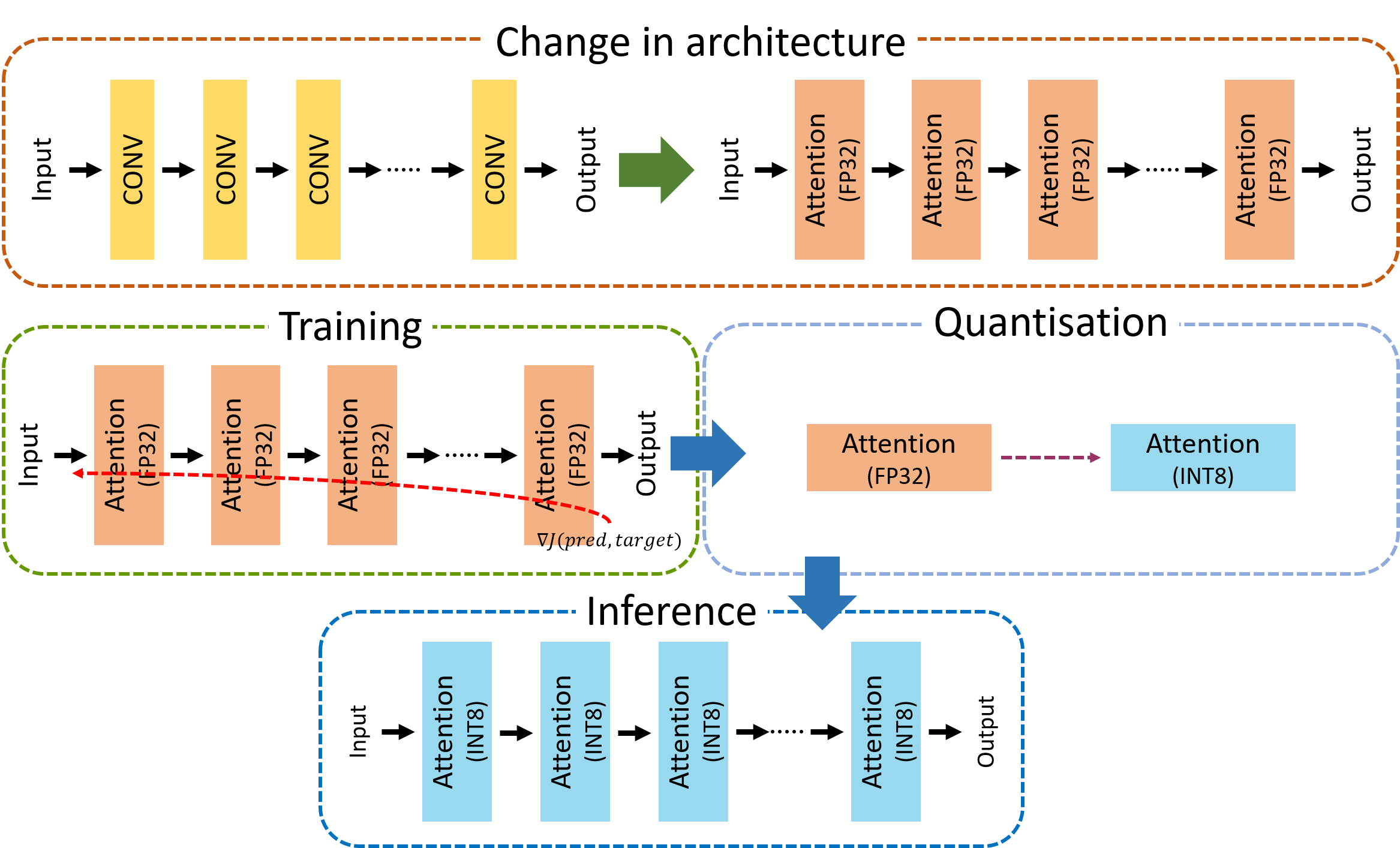}
    \caption{\textbf{Overview of the proposed method.} Convolutional layers in deep neural network architectures are replaced with self-attention layers and networks with parameters at FP32 precision are trained till convergence. To optimise the model for storage and faster inference, the network parameters are quantised without loss in performance.}
    \label{fig:workflow}
\end{figure}

\noindent\textbf{Our Approach:}
Inspired by the success of \cite{ramachandran2019stand} in natural image classification tasks, we propose the design of a new class of networks for medical image classification and segmentation, in which we replace the convolution layers with self-attention layers. Furthermore, we optimise the networks for inference by quantising the parameters thereby decreasing energy consumption. To the best of our knowledge, a quantised fully self-attentive network for classification and segmentation of medical images and comparison with its convolutional counterparts has not been attempted so far. Schematic overview of the proposed method is illustrated in Fig.~\ref{fig:workflow}.

\section{Method}
\subsection{Stand-alone self-attention}

Attention was introduced by \cite{bahdanau2014neural} for a neural machine translation model. Attention modules can learn to focus on essential regions within a context, making it an important component of neural networks. Self-attention \cite{vaswani2017attention} is defined as attention applied to a single context instead of across multiple contexts; that is, \textit{Key}, \textit{Query} and \textit{Values} are derived from the same context. \cite{ramachandran2019stand} introduced the stand-alone self-attention layer, which can replace convolutions to construct a fully attentional model. Motivated by the initial success of \cite{ramachandran2019stand} in natural images, we explore the feasibility of using such modules in the proposed class of networks for medical image analysis. 

To compute attention for each pixel $\mathbf{x}_{i,j} \in \mathbb{R}^{C_{in} \times 1 \times 1}$ in an image or an activation map, local regions with spatial extent $h\times w$ around $\mathbf{x}_{i,j}$ are used to derive the \textit{keys} and \textit{values}. Learned linear transformations are performed on $\mathbf{x}_{i,j}$ and its local regions to obtain \textit{query} ($\mathbf{Q}$), \textit{keys} ($\mathbf{K}$) and \textit{values} ($\mathbf{V}$) as

\begin{equation}
    \mathbf{Q} = \mathbf{W_{Q}}\mathit{\mathbf{x}_{i,j}}
    \label{eq:query}
\end{equation}
\begin{equation}
   \mathbf{K} = \mathbf{W_{K}}\mathit{\mathbf{x}_{h,w}}
    \label{eq:key}
\end{equation}
\begin{equation}
    \mathbf{V} = \mathbf{W_{V}}\mathit{\mathbf{x}_{h,w}}
    \label{eq:value}
\end{equation}
\noindent where $\mathbf{W_{Q}} \in \mathbb{R}^{C_{out} \times C_{in}}$, $\mathbf{W_{K}} \in \mathbb{R}^{C_{out} \times C_{in}}$ and $\mathbf{W_{V}} \in \mathbb{R}^{C_{out} \times C_{in}}$ are learnable transformation matrices and $\mathbf{x}_{h,w}\in \mathbb{R}^{C_{in}\times h \times w}$ is the local region centered at $\mathbf{x}_{i,j}$.

\begin{figure}
    \centering
    \includegraphics[width=0.9\linewidth]{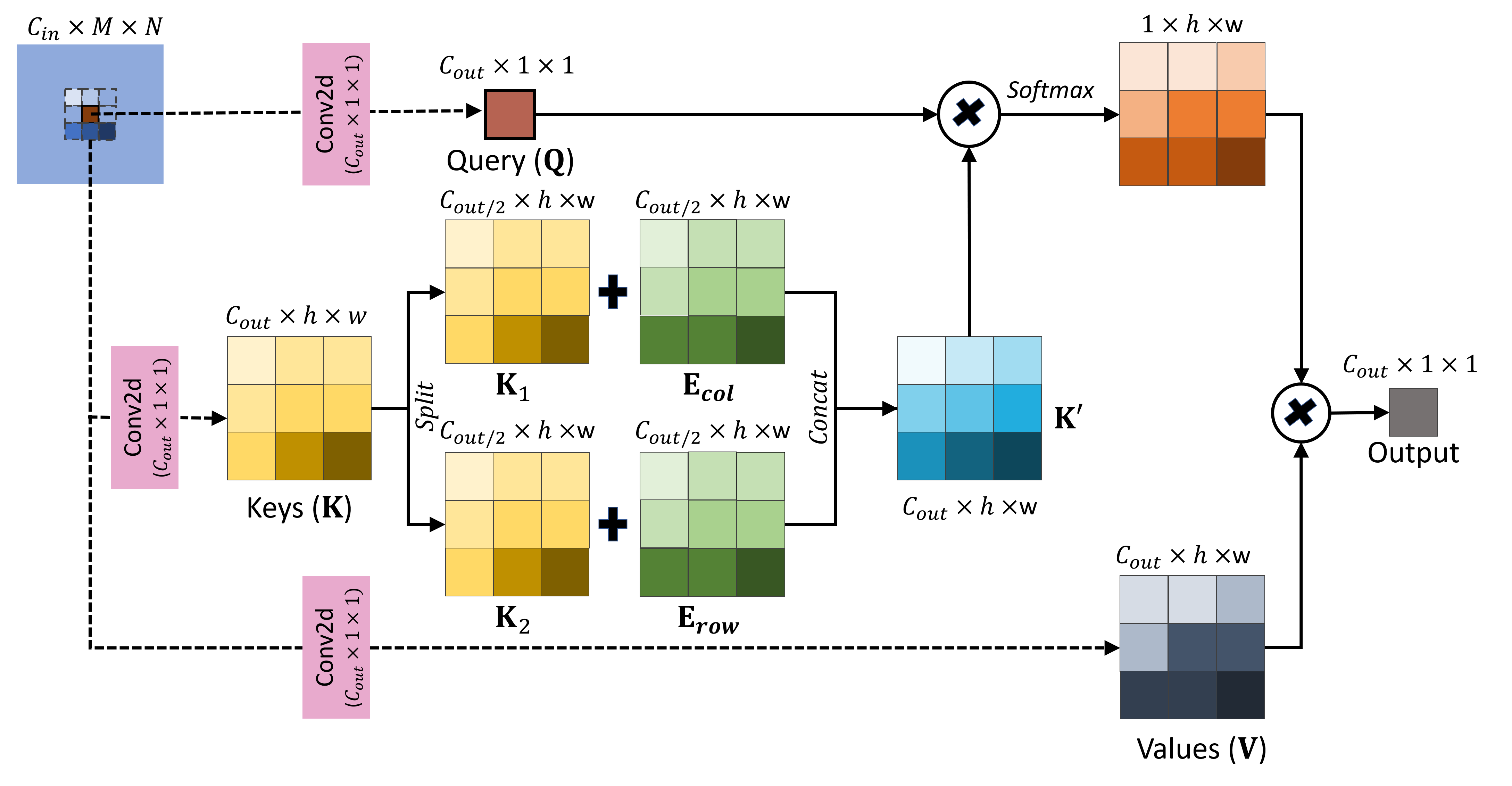}
    \caption{\textbf{Self-attention mechanism with local context.} Operations are performed on a per-pixel basis to compute attention as shown in the figure. Linear transformations for obtaining query, keys and values are implemented using 2D convolution (\textit{Conv2d}) operation. The learnt relative positional embedding are added to the keys to incorporate the inter-pixel relationships within the local context.}
    \label{fig:self_atention}
\end{figure}

Self-attention on its own does not encode any positional information, which makes it permutation equivariant. Relative positional embedding \cite{shaw2018self} as used in \cite{ramachandran2019stand} are incorporated into the attention module. The keys $\mathbf{K}\in\mathbb{\rm I\!R^{C_{out}}}\times h \times w$ are split into $\mathbf{K}_1, \mathbf{K}_2\in\mathbb{\rm I\!R^{C_{out/2}\times h \times w}}$ each and column offset $\mathbf{E}_{col}$ and row offset $\mathbf{E}_{row}$ of the positional embedding are added to these separately. After this, we concatenate $\mathbf{K}_1, \mathbf{K}_2$ to obtain a new key ($\mathbf{K}’\in \mathbb{\rm I\!R^{C_{out}\times h \times w}}$) which contains the relative spatial information of pixels in the local region of size $h \times w$. Thus, the relative spatial attention for a pixel $x_{ij}$ is mathematically defined as in Eq.~\ref{eq:attention} and is graphically illustrated in Figure~\ref{fig:self_atention}.

\begin{equation}
    \mathbf{y_{i,j}} = \sum_{\{u,v\} \in N_{h,w}(i,j)}\mathit{softmax}_{u,v}(\mathbf{Q}_{i,j}^\top \mathbf{K}_{u,v})\mathbf{V}_{u,v} 
    \label{eq:attention}
\end{equation}
\noindent where $N_{h,w}(i,j)$ is the neighbourhood of size $h \times w$ centered at $(i,j)$.

We use these attention blocks instead of 2D convolutional blocks in our networks. During training, all the weights and activations are represented and stored with a precision of FP32. The parameters are quantised to INT8 precision for inference. 

\subsection{Quantisation of network parameters}
We perform quantisation using the FBGEMM (FaceBook GEneral Matrix Multiplication) \cite{fbgemm} backend of PyTorch for x86 CPUs, which is based on the quantisation scheme proposed by \cite{jacob2018quantization}. In order to be able to perform all the arithmetic operations using integer arithmetic operations on quantised values, we require the quantisation scheme to be an affine mapping of integers $q$ to real numbers $r$ as

\begin{equation}
    r = S(q-Z)
    \label{eq:quantisation}
\end{equation}

\noindent where $S$ and $Z$ are quantisation parameters. We have employed a post-training $8$-bit quantisation of all the weights and operations for our proposed model.

\subsection{Network architecture}
\begin{figure}[h]
    \centering
    \subfigure[Classification network (SaDNN-cls)]{ \includegraphics[width=0.75\linewidth]{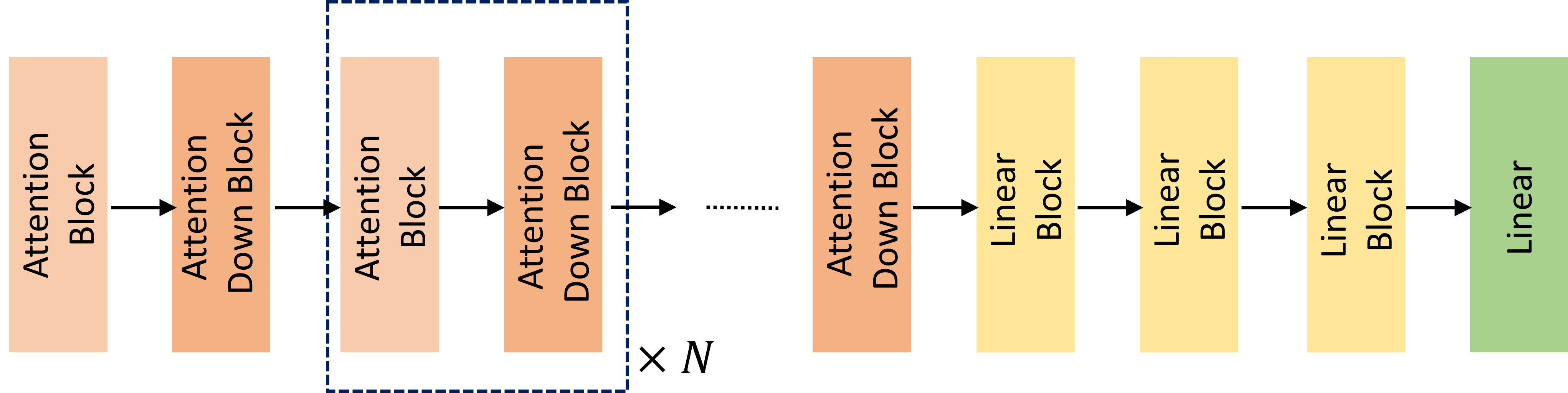}\label{fig:network_cls}}
    
    \subfigure[Segmentation network (SaDNN-seg)]{ \includegraphics[width=0.9\linewidth]{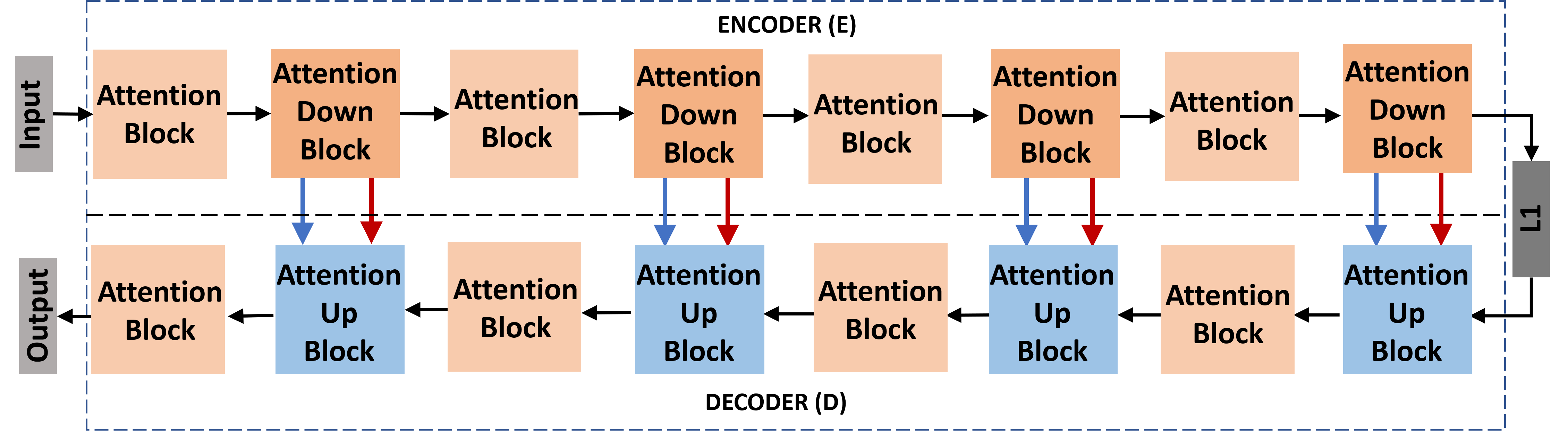}\label{fig:network_seg}}
    
    \subfigure[Network components]{ \includegraphics[width=0.7\linewidth]{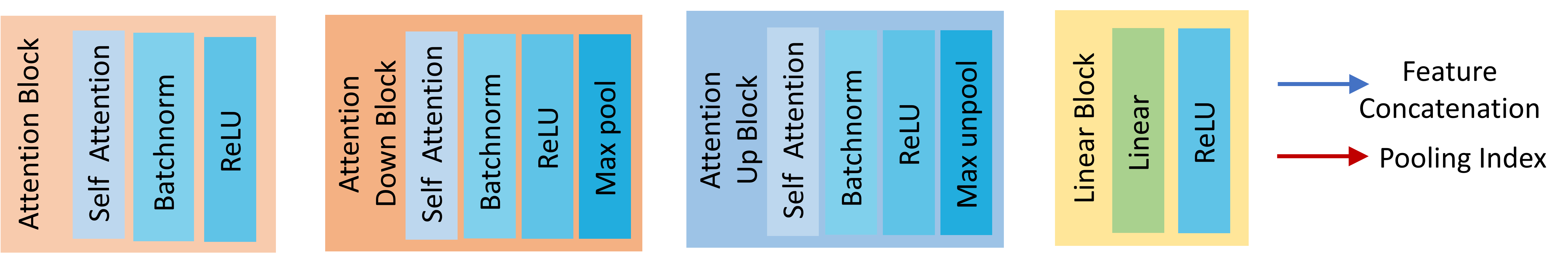}\label{fig:network_comps}}
    \caption{\textbf{Architecture of the proposed Self-attentive Deep Neural Networks (SaDNN).} Detailed architecture of the networks for classification and segmentation are shown in (a) and (b) respectively. Components of the various blocks in these networks are detailed in (c).}
    \label{fig:network}
\end{figure}

\textbf{Classification:} The architecture of the proposed classification network is illustrated in Fig.~\ref{fig:network_cls} with the details of the constituent modules in Fig.~\ref{fig:network_comps}. The network consists of a series of alternating attention blocks and attention down blocks followed by fully-connected linear layers. The feature maps are downsampled using the max-pooling operation. The size of the output linear layer is equal to the number of target classes. The network is trained to perform multi-label classification using a binary cross-entropy loss.

\noindent\textbf{Segmentation:} The proposed segmentation network has a fully attention-based encoder-decoder architecture as shown in Fig.~\ref{fig:network_seg}. The encoder unit consists of stand-alone self-attention blocks with ReLU activation and max-pooling operations with the number of feature maps increasing progressively with each attention block. The decoder consists of attention blocks and max-unpooling operations. The size of activation maps of the decoder matches with the corresponding layer in the encoder. The unpooling operations are performed using the indices transferred from the pooling layers in the encoder. To prevent the loss of subtle information, we employ activation concatenation in the decoder, similar to UNet \cite{ronneberger2015u}. The network is trained using soft dice loss \cite{7785132}. 

\section{Experiments}
\subsection{Datasets}

\textbf{Classification:} To evaluate the performance of the fully self-attentive network (SaDNN-cls) on classification tasks, we have used the NIH Chest X-ray dataset of $14$ Common Thorax Disease \cite{wang2017chestx}. The dataset comprises $112,120$ frontal-view X-ray images of $30,805$ patients with fourteen disease labels. These disease classes can co-occur in an image; therefore, the classification problem is formulated as multi-label classification. The train, validation and test split provided in the dataset was used for the experiments.

\noindent \textbf{Segmentation:} A subset of the medical segmentation decathlon dataset \cite{antonelli2021medical} is used to evaluate the performance of the proposed fully-attentive network (SaDNN-seg) for liver segmentation. Out of the $131$ ground truth paired $3$D CT volumes-Ground truth pairs available in the dataset, $80$ per cent were randomly chosen for training, and the remaining $20$ per cent were used for testing.

\subsection{Implementation Details}
\noindent \textbf{Training:} The proposed models were trained using an Adam Optimiser \cite{kingma2014adam} with a learning rate of $1\times 10^{-4}$. The models for classification task were trained for $15$ epochs and the models for segmentation were trained for $25$ epochs.

\noindent \textbf{Baselines:} Performance of the proposed quantised self-attention network for the classification task is compared with ResNet-18, ResNet-50 and their $8-$bit quantised versions q-ResNet-18, and q-ResNet-50. To assess the performance of the segmentation network, we chose a modified UNet\cite{ronneberger2015u} (UNet-small) and SUMNet\cite{nandamuri2019sumnet} architecture trained on the same dataset split and their quantised versions q-UNet-small and q-SUMNet as baselines. 

\noindent \textbf{System specifications}: All networks were trained on a high-performance server with a NVIDIA $V100$ GPU, $x86_64$ Intel(R) Xeon(R) Silver $4110$ CPU @ $2.10GHz$, $96$ GB RAM and $1$ TB HDD running on Ubuntu $18.01.1$ LTS OS. The inference of quantised models was also performed on the same class of CPUs.

\section{Results and Discussions}
\subsection{Qualitative Analysis}
\begin{figure}
    \centering
    \subfigure[Sample 1]{\includegraphics[width=0.22\linewidth]{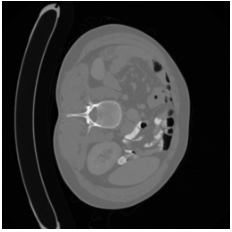}}
    \subfigure[Ground Truth]{\includegraphics[width=0.22\linewidth]{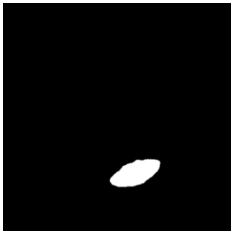}}
    \subfigure[SaDNN-seg]{\includegraphics[width=0.22\linewidth]{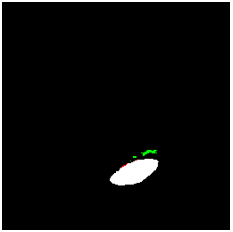}}
    \subfigure[q-SaDNN-seg]{\includegraphics[width=0.22\linewidth]{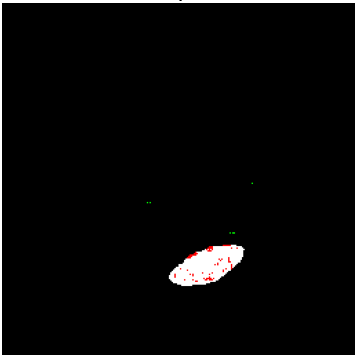}}
    
    \subfigure[Sample 2]{\includegraphics[width=0.22\linewidth]{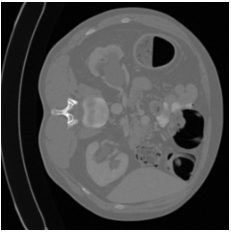}}
    \subfigure[Ground Truth]{\includegraphics[width=0.22\linewidth]{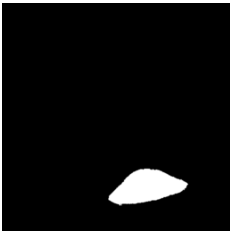}}
    \subfigure[SaDNN-seg]{\includegraphics[width=0.22\linewidth]{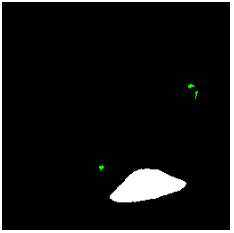}}
    \subfigure[q-SaDNN-seg]{\includegraphics[width=0.22\linewidth]{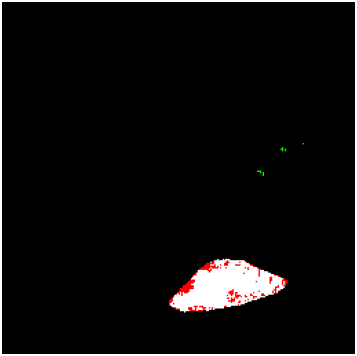}}
   
    \caption{\textbf{Comparison of segmentation predictions.} Figure shows sample input CT images in (a) and (e) with the corresponding ground truths of liver in (b) and (f) respectively. Segmentation map as predicted by SaDNN-seg, with the over-segmented region marked in green and under-segmented region marked in red are presented in (c) and (g) for the two sample images. Similar visualisation of segmentation by the proposed q-SaDNN-seg are presented in (d) and (h).}
    \label{fig:pred_vis}
\end{figure}

visualisation of predictions of the proposed q-SaDNN-seg network and its unquantised version SaDNN-seg are presented in Fig.~\ref{fig:pred_vis}. Over-segmented regions in the predicted segmentation maps are marked in green, under-segmented regions are marked in red and correctly segmented region is shown in white. We observe that the tendency of the original unquantised network SaDNN-seg to over-segment is significantly reduced post quantisation. However, the quantisation of network parameters causes the q-SaDNN-seg to under-segment the target organ. This is reflected in the slightly lower Dice coefficient (DSC) of the proposed model as seen in Table~\ref{tab:quant_segmentation}. 

\vspace{-1cm}

\subsection{Quantitative Analysis}
The performance of the proposed quantised fully self-attentive network and baselines for multi-label classification task is reported in terms of accuracy in Table~\ref{tab:quant_classification}. It can be observed that the proposed network can achieve performance slightly better than the existing deep residual convolutional neural networks. Table~\ref{tab:quant_segmentation} shows the comparison of the proposed segmentation network with the baselines in terms of DSC. The proposed quantised network performs almost as good as the quantised versions of the baseline convolutional neural networks. 

\begin{figure}
\CenterFloatBoxes
\begin{floatrow}
\ttabbox
  {\begin{tabular}{|c|c|}
\hline
Model              & Accuracy      \\ \hline
ResNet-18          & 0.89          \\ \hline
q-ResNet-18        & 0.88          \\ \hline
ResNet-50          & 0.84          \\ \hline
q-ResNet-50        & 0.83          \\ \hline
SaDNN-cls (ours)   & \textbf{0.90} \\ \hline
q-SaDNN-cls (ours) & 0.89          \\ \hline
\end{tabular}}
  {\caption{Evaluation of classification}\label{tab:quant_classification}}
\killfloatstyle
\ttabbox
  {\begin{tabular}{|c|c|}
\hline
Model              & DSC    \\ \hline
UNet-small               & 0.88          \\ \hline
q-UNet-small             & 0.88              \\ \hline
SUMNet             & 0.89          \\ \hline
q-SUMNet           & 0.89              \\ \hline
SaDNN-seg (ours)   & \textbf{0.88} \\ \hline
q-SaDNN-seg (ours) & 0.85     \\ \hline
\end{tabular}
  }
  {\caption{Evaluation of segmentation}\label{tab:quant_segmentation}}
\end{floatrow}
\end{figure}



\vspace{-0.5cm}

\subsection{Computational Analysis}

The DNNs used for the experiments exhibited superior classification and segmentation performance in terms of quantitative metrics, but they require a considerable amount of computations and memory access operations to be performed. Deploying a framework which needs excessive computations to be performed results in large energy consumption, which is not feasible in diverse resource-constrained scenarios. Therefore, it is key to have an energy-efficient model without degradation in performance. A rough estimate of energy cost per operation in $45nm$ $ 0.9V$ IC design can be calculated using Table \ref{tab:energy_ref} presented in \cite{horowitz20141,wu2018training,park2020cenna}. 

\begin{table}[]
\caption{Approximate energy cost in $45nm$ $0.9V$ for different multiplication and addition operations}
\label{tab:energy_ref}
\begin{tabular}{|l|ll|}
\hline
\multirow{2}{*}{\textbf{Operation}} & \multicolumn{2}{l|}{\textbf{Energy (pJ)}}        \\ \cline{2-3} 
                                    & \multicolumn{1}{l|}{\textbf{MUL}} & \textbf{ADD} \\ \hline
8-bit INT                           & \multicolumn{1}{l|}{0.2 pJ}       & 0.03 pJ      \\ \hline
16-bit FP                           & \multicolumn{1}{l|}{1.1 pJ}       & 0.40 pJ      \\ \hline
32-bit FP                           & \multicolumn{1}{l|}{3.7 pJ}       & 0.90 pJ      \\ \hline
\end{tabular}
\end{table}


The number of multiplication and addition operations in a standalone self-attention layer \cite{vaswani2021scaling} can be calculated as
\begin{equation}
Ops_{mul} = Ops_{add} = 2b^2c
\label{eq:10}
\end{equation}
\noindent where $b$ is the block (local region) size and $c$ is the number of channels.

\begin{figure}[h]
    \centering
    \subfigure[Classification ]{\includegraphics[width=0.45\linewidth]{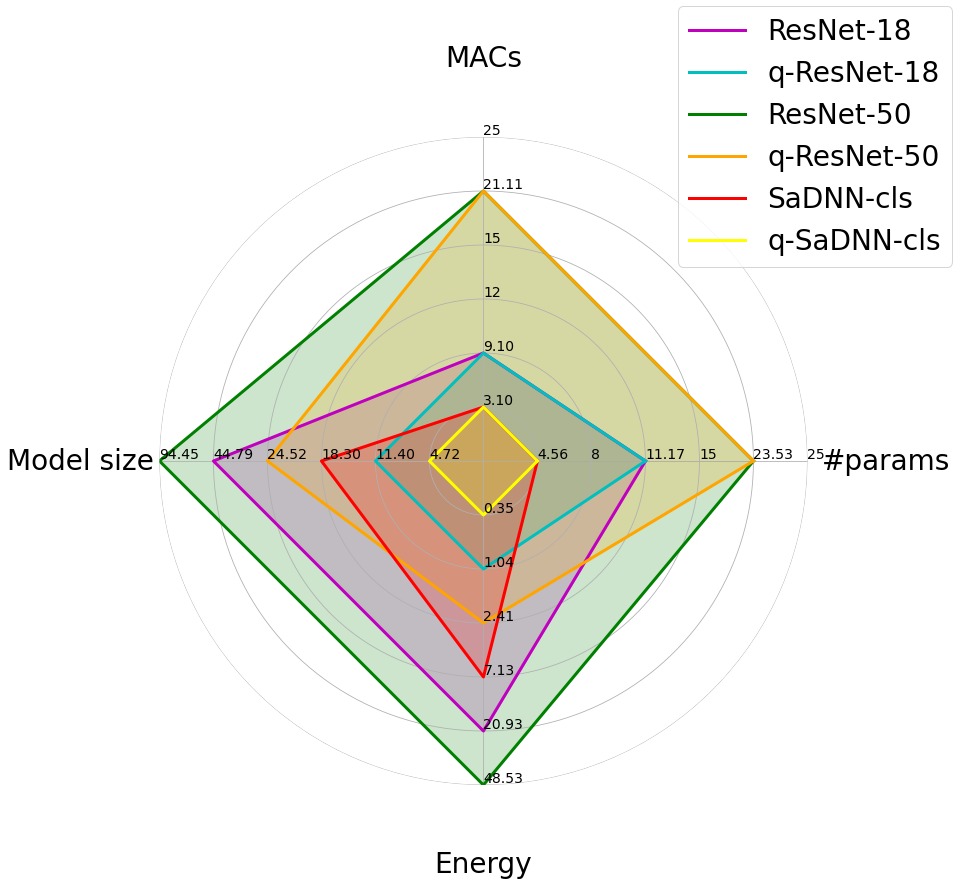}}
    \subfigure[Segmentation]{\includegraphics[width=0.45\linewidth]{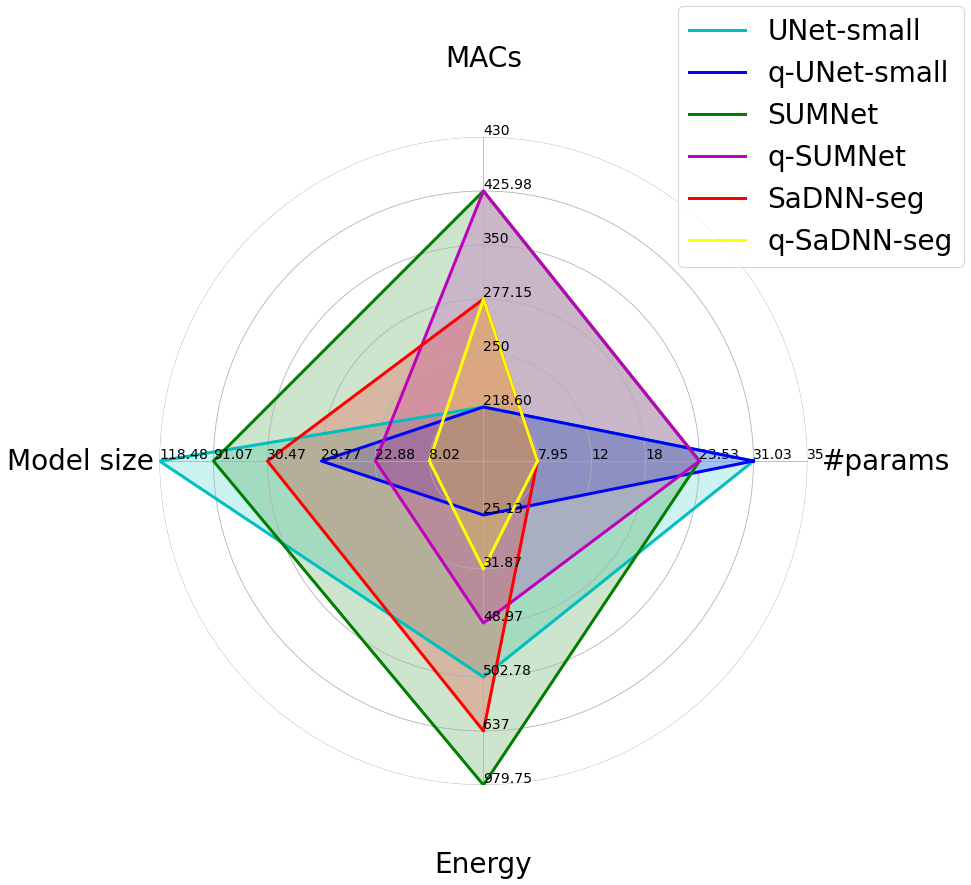}}
    \caption{\textbf{Graphical comparison of proposed networks.} Figure shows radar chart based comparison of proposed (a) classification network and (b) segmentation network in terms of number of parameters, MACs, model size and energy. The model with the least area within the plot is the best one.}
    \label{fig:radar_plot}
\end{figure}

The total number of parameters, MACs, energy consumed during forward pass and model size of the proposed q-SaDNN-cls and q-SaDNN-seg networks are reported in Table~\ref{tab:compute_classification} and Table~\ref{tab:compute_segmentation} with graphical comparisons in Fig.~\ref{fig:radar_plot}. Models with the least area in the radar charts are more efficient. The proposed q-SaDNN-cls network is $58.59\%$ smaller than quantised ResNet-18 and $80.75\%$ smaller than quantised ResNet-50 in terms of model size. In terms of total MAC units, the the propsed networks have $65.93\%$ fewer MACs than ResNet-18, $85.32\%$ fewer than ResNet-50. Similarly, in terms of the total trainable parameters, the proposed networks have $59.17\%$ lesser parameters than ResNet-18 and $80.62\%$ lesser than ResNet-50.

\begin{table}[h]
\caption{Comparison of classification networks}\label{tab:compute_classification}
\begin{tabular}{|c|c|c|c|c|}
\hline
Model              & \#Params      & MACs & Model size & Energy \\ \hline
ResNet-18          & 11.17 M         &   9.10 G    & 44.79 MB   &   20.93 J       \\ \hline
q-ResNet-18        & 11.17 M         &   9.10 G    & 11.40 MB   &   1.04 J       \\ \hline
ResNet-50          & 23.53 M         &   21.11 G    & 94.45 MB   &  48.53 J       \\ \hline
q-ResNet-50        & 23.53 M         &   21.11 G    & 24.52 MB   &   2.41 J       \\ \hline
\textbf{SaDNN-cls}    & \textbf{4.56 M} &   3.10 G    & 18.30 MB    &  7.13 J       \\ \hline
\textbf{q-SaDNN-cls}  & \textbf{4.56 M} &   3.10 G    & \textbf{4.72 MB}  & 0.35 J   \\ \hline
\end{tabular}
\end{table}


Similar improvement in efficiency of computing can be observed in the case of segmentation as well. The segmentation network q-SaDNN-seg is $73.06\%$ smaller than q-UNet-small and $64.94\%$ smaller than q-SUMNet in terms of model size. In terms of total MAC units, the q-SaDNN-seg has $34.94\%$ fewer than SUMNet. In terms of the trainable parameters, q-SaDNN-seg has $74.37\%$ lesser parameters than UNet-small and $66.21\%$ lesser than SUMNet. It is to be noted that the proposed models are superior in terms energy consumption as well.

\begin{table}[!h]
\caption{Comparison of segmentation networks}\label{tab:compute_segmentation}
    \centering
    \begin{tabular}{|c|c|c|c|c|}
    \hline
    Model              & \#Params   & MACs & Model size & Energy       \\ \hline
    UNet-small               & 31.03 M         &   218.60 G    & 118.48 MB  & 502.78 J     \\ \hline
    q-UNet-small             & 31.03 M         &   218.60 G    & 29.77 MB   & 25.13 J     \\ \hline
    SUMNet             & 23.53 M         &   425.98 G    & 91.07 MB   & 979.75 J     \\ \hline
    q-SUMNet           & 23.53 M         &   425.98 G    & 22.88 MB  &  48.97 J     \\ \hline
    \textbf{SaDNN-seg}   & \textbf{7.95 M} &   277.15 G    & 30.47 MB  &  637 J     \\ \hline
    \textbf{q-SaDNN-seg} & \textbf{7.95 M} &   277.15 G    & \textbf{8.02 MB}& 31.87 J\\ \hline
    \end{tabular}
\end{table}

\begin{figure}[h]
    \centering
    \subfigure[Test image]{\includegraphics[width=0.3\linewidth]{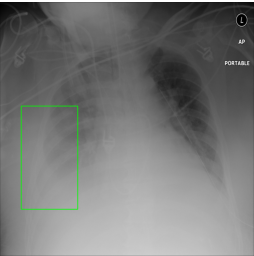}}
    \subfigure[Saliency map]{\includegraphics[width=0.3\linewidth]{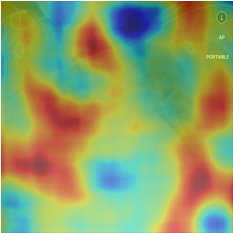}}
    \caption{Figure shows (a) a sample image from the test set used in our experiments with the clinically relevant region as provided in the dataset marked in green and (b) saliency map of q-SaDNN-cls. Regions shown in red in the saliency map are perceived as most important and those in blue to be least important by the network during prediction.}
    \label{fig:rise_maps}
\end{figure}
 
\subsection{Analysis of clinical relevance}
Validating the results of the model with respect to clinically relevant information to provide some explanations for the decision made by the model is an important factor that determines trustability. The clinically relevant region provided in the NIH Chest X-ray dataset as marked by a radiologist and the
saliency map based explanation generated using RISE\cite{petsiuk2018rise} for the proposed quantised self-attention deep neural network for classification are shown in Fig.~\ref{fig:rise_maps}. It can be observed that the proposed model focuses on the clinically relevant region while making the decision.

\section{Conclusion}
We proposed a class of quantised self-attentive neural networks which can be used for medical image classification and segmentation. In these networks, convolutional layers are replaced with attention layers which have fewer learnable parameters. Computation of attention while considering a small local region surrounding a pixel prevents degradation of performance despite the absence of local feature extraction which is typically performed in a CNN. We show that our energy efficient method achieves performance at par with the commonly used CNNs with fewer number of parameters and model size. These attributes make our proposed models affordable and easy to adopt in resource constrained settings.

%
%
%
\bibliographystyle{splncs04}
\bibliography{8}

\begin{thebibliography}{10}
\providecommand{\url}[1]{\texttt{#1}}
\providecommand{\urlprefix}{URL }
\providecommand{\doi}[1]{https://doi.org/#1}

\bibitem{aji2020compressing}
Aji, A.F., Heafield, K.: Compressing neural machine translation models with
  4-bit precision. In: Workshop Neural Gen. Transl. pp. 35--42 (2020)

\bibitem{antonelli2021medical}
Antonelli, M., Reinke, A., Bakas, S., Farahani, K., Landman, B.A., Litjens, G.,
  Menze, B., Ronneberger, O., Summers, R.M., van Ginneken, B., et~al.: The
  medical segmentation decathlon. arXiv preprint arXiv:2106.05735  (2021)

\bibitem{bahdanau2014neural}
Bahdanau, D., Cho, K., Bengio, Y.: Neural machine translation by jointly
  learning to align and translate. arXiv preprint arXiv:1409.0473  (2014)

\bibitem{chakravartyradiologist}
Chakravarty, A., Ghosh, N., Sheet, D., Sarkar, T., Sethuraman, R.: Radiologist
  validated systematic search over deep neural networks for screening
  musculoskeletal radiographs  (2019)

\bibitem{hatamizadeh2022unetr}
Hatamizadeh, A., Tang, Y., Nath, V., Yang, D., Myronenko, A., Landman, B.,
  Roth, H.R., Xu, D.: Unetr: Transformers for 3d medical image segmentation.
  In: .IEEE/CVF Winter Conf. App. Comp. Vis. pp. 574--584 (2022)

\bibitem{he2016deep}
He, K., Zhang, X., Ren, S., Sun, J.: Deep residual learning for image
  recognition. In: Proc. IEEE Conf. Comp. Vis. Patt. Recog. pp. 770--778 (2016)

\bibitem{horowitz20141}
Horowitz, M.: 1.1 computing's energy problem (and what we can do about it). In:
  IEEE Int. Solid-State Circuits Conf. Digest of Technical Papers. pp. 10--14
  (2014)

\bibitem{huang2017densely}
Huang, G., Liu, Z., Van Der~Maaten, L., Weinberger, K.Q.: Densely connected
  convolutional networks. In: Proc. IEEE Conf. Comp. Vis. Patt. Recog. pp.
  4700--4708 (2017)

\bibitem{jacob2018quantization}
Jacob, B., Kligys, S., Chen, B., Zhu, M., Tang, M., Howard, A., Adam, H.,
  Kalenichenko, D.: Quantization and training of neural networks for efficient
  integer-arithmetic-only inference. In: IEEE Conf. Comp. Vis. Patt. Recog. pp.
  2704--2713 (2018)

\bibitem{fbgemm}
Khudia, D., Huang, J., Basu, P., Deng, S., Liu, H., Park, J., Smelyanskiy, M.:
  Fbgemm: Enabling high-performance low-precision deep learning inference.
  arXiv preprint arXiv:2101.05615  (2021)

\bibitem{kingma2014adam}
Kingma, D.P., Ba, J.: Adam: A method for stochastic optimization. arXiv
  preprint arXiv:1412.6980  (2014)

\bibitem{7785132}
Milletari, F., Navab, N., Ahmadi, S.A.: V-net: Fully convolutional neural
  networks for volumetric medical image segmentation. In: 2016 Fourth
  International Conference on 3D Vision (3DV). pp. 565--571 (2016).
  \doi{10.1109/3DV.2016.79}

\bibitem{nandamuri2019sumnet}
Nandamuri, S., China, D., Mitra, P., Sheet, D.: Sumnet: Fully convolutional
  model for fast segmentation of anatomical structures in ultrasound volumes.
  In: IEEE Int. Symp. Biomed. Imag. pp. 1729--1732 (2019)

\bibitem{park2020cenna}
Park, S.S., Chung, K.S.: Cenna: Cost-effective neural network accelerator.
  Electronics  \textbf{9}(1), ~134 (2020)

\bibitem{paupamah2020quantisation}
Paupamah, K., James, S., Klein, R.: Quantisation and pruning for neural network
  compression and regularisation. In: Int.l SAUPEC/RobMech/PRASA Conf.
  pp.~1--6. IEEE (2020)

\bibitem{petsiuk2018rise}
Petsiuk, V., Das, A., Saenko, K.: Rise: Randomized input sampling for
  explanation of black-box models. Brit. Machine Vis. Conf.  (2018)

\bibitem{ramachandran2019stand}
Ramachandran, P., Parmar, N., Vaswani, A., Bello, I., Levskaya, A., Shlens, J.:
  Stand-alone self-attention in vision models. arXiv preprint arXiv:1906.05909
  (2019)

\bibitem{ronneberger2015u}
Ronneberger, O., Fischer, P., Brox, T.: U-net: Convolutional networks for
  biomedical image segmentation. In: Int. Conf. Med. image Comput.
  Comp.r-assist. Interv. pp. 234--241. Springer (2015)

\bibitem{shaw2018self}
Shaw, P., Uszkoreit, J., Vaswani, A.: Self-attention with relative position
  representations. arXiv preprint arXiv:1803.02155  (2018)

\bibitem{vaswani2021scaling}
Vaswani, A., Ramachandran, P., Srinivas, A., Parmar, N., Hechtman, B., Shlens,
  J.: Scaling local self-attention for parameter efficient visual backbones.
  In: Proceedings of the IEEE/CVF Conference on Computer Vision and Pattern
  Recognition. pp. 12894--12904 (2021)

\bibitem{vaswani2017attention}
Vaswani, A., Shazeer, N., Parmar, N., Uszkoreit, J., Jones, L., Gomez, A.N.,
  Kaiser, {\L}., Polosukhin, I.: Attention is all you need. In: Adv. Neural
  Info. Process. Sys. pp. 5998--6008 (2017)

\bibitem{wang2017chestx}
Wang, X., Peng, Y., Lu, L., Lu, Z., Bagheri, M., Summers, R.M.: Chestx-ray8:
  Hospital-scale chest x-ray database and benchmarks on weakly-supervised
  classification and localization of common thorax diseases. In: IEEE Conf.
  Comp. Vis. Patt. Recog. pp. 2097--2106 (2017)

\bibitem{wu2018training}
Wu, S., Li, G., Chen, F., Shi, L.: Training and inference with integers in deep
  neural networks. arXiv preprint arXiv:1802.04680  (2018)

\bibitem{xu2021mixed}
Xu, J., Yu, J., Hu, S., Liu, X., Meng, H.M.: Mixed precision low-bit
  quantisation of neural network language models for speech recognition.
  IEEE/ACM Trans. Audio, Speech, Lang. Process.  (2021)

\bibitem{yang2020learning}
Yang, F., Yang, H., Fu, J., Lu, H., Guo, B.: Learning texture transformer
  network for image super-resolution. In: IEEE/CVF Conf. Comp. Vis. Patt.
  Recog. pp. 5791--5800 (2020)

\bibitem{ye2019cross}
Ye, L., Rochan, M., Liu, Z., Wang, Y.: Cross-modal self-attention network for
  referring image segmentation. In: IEEE/CVF Conf. Comp. Vis. Patt. Recog. pp.
  10502--10511 (2019)

\bibitem{zhu2020deformable}
Zhu, X., Su, W., Lu, L., Li, B., Wang, X., Dai, J.: Deformable detr: Deformable
  transformers for end-to-end object detection. arXiv preprint arXiv:2010.04159
   (2020)

\end{thebibliography}
%

\end{document}